\documentclass[final]{cvpr}

\usepackage[accsupp]{axessibility}
\usepackage{times}
\usepackage{epsfig}
\usepackage{graphicx}
\usepackage{amsmath}
\usepackage{amssymb}
\usepackage{float}
\usepackage{mathrsfs}
\usepackage{multirow}

\usepackage{array}
\usepackage{bm}
\usepackage{multirow}
\usepackage{color}
\usepackage{bm}
\usepackage{makecell}
\usepackage{colortbl}
\definecolor{mygray}{gray}{.9}
\usepackage{wrapfig}

\usepackage[pagebackref=true,breaklinks=true,colorlinks]{hyperref}

\def\cvprPaperID{14922} 
\def\confName{CVPR}
\def\confYear{2024}
\graphicspath{{./figs/}{./figs/result/}}

\begin{document}
	
	\title{Progressive Semantic-Guided Vision Transformer for  Zero-Shot Learning}
	
	\author{Shiming Chen$^{1}$, Wenjin Hou$^{2}$, Salman Khan$^{1,3}$, Fahad Shahbaz Khan$^{1,4}$\\
		$^{1}$Mohamed bin Zayed University of AI \quad
		$^{2}$Huazhong University of Science and Technology \\
		$^{3}$Australian National University \quad
		$^{4}$Linköping  University  \\
		 \quad
		{\tt\small \{shimingchen, houwj17\}@gmail.com \quad \{salman.khan, fahad.khan\}@mbzuai.ac.ae}
	}
	
	\maketitle

	\begin{abstract}
		{Zero-shot learning (ZSL) recognizes the unseen classes by conducting visual-semantic interactions to transfer semantic knowledge from seen classes to unseen ones, supported by semantic information (\textit{e.g.}, attributes). However, existing ZSL methods simply extract visual features using a pre-trained network backbone (\textit{i.e.}, CNN or ViT), which fail to learn matched visual-semantic correspondences for representing semantic-related visual features as lacking of the guidance of semantic information, resulting in undesirable visual-semantic interactions. To tackle this issue, we propose a progressive semantic-guided vision transformer for zero-shot learning (dubbed ZSLViT). ZSLViT mainly considers two properties in the whole network: i) discover the semantic-related visual representations explicitly, and ii) discard the semantic-unrelated visual information. Specifically, we first introduce semantic-embedded token learning to improve the visual-semantic correspondences via semantic enhancement and discover the semantic-related visual tokens explicitly with semantic-guided token attention. Then, we fuse low semantic-visual correspondence visual tokens to discard the semantic-unrelated visual information for visual enhancement. These two operations are integrated into various encoders to progressively learn semantic-related visual representations for accurate visual-semantic interactions in ZSL. The extensive experiments show that our ZSLViT achieves significant performance gains on three popular benchmark datasets, \textit{i.e.}, CUB, SUN, and AWA2.  Codes are available at: \url{https://github.com/shiming-chen/ZSLViT}.
		}
	\end{abstract}
	\section{Introduction}\label{Sec1}

	Zero-shot learning (ZSL), aiming to recognize unseen classes by exploiting the intrinsic semantic relatedness between seen and unseen categories during
	training \cite{Lampert2009LearningTD, Larochelle2008ZerodataLO, Palatucci2009ZeroshotLW,Zhang2023TowardsRZ}, has achieved significant progress. Inspired by the way humans learn unknown concepts, semantic information (e.g., attributes \cite{Lampert2014AttributeBasedCF}) shared by seen and unseen classes is employed to support knowledge transfer from seen classes to unseen ones. Targeting this goal, ZSL conducts effective visual-semantic interactions between visual and semantic spaces to align them. For example, discovering the semantic representations in visual spaces and matching them with the semantic information. As such, exploring the shared semantic knowledge between the visual and semantic spaces is essential.

	\begin{figure*}[t]
		\begin{center}
			\includegraphics[width=17cm,height=6.2cm]{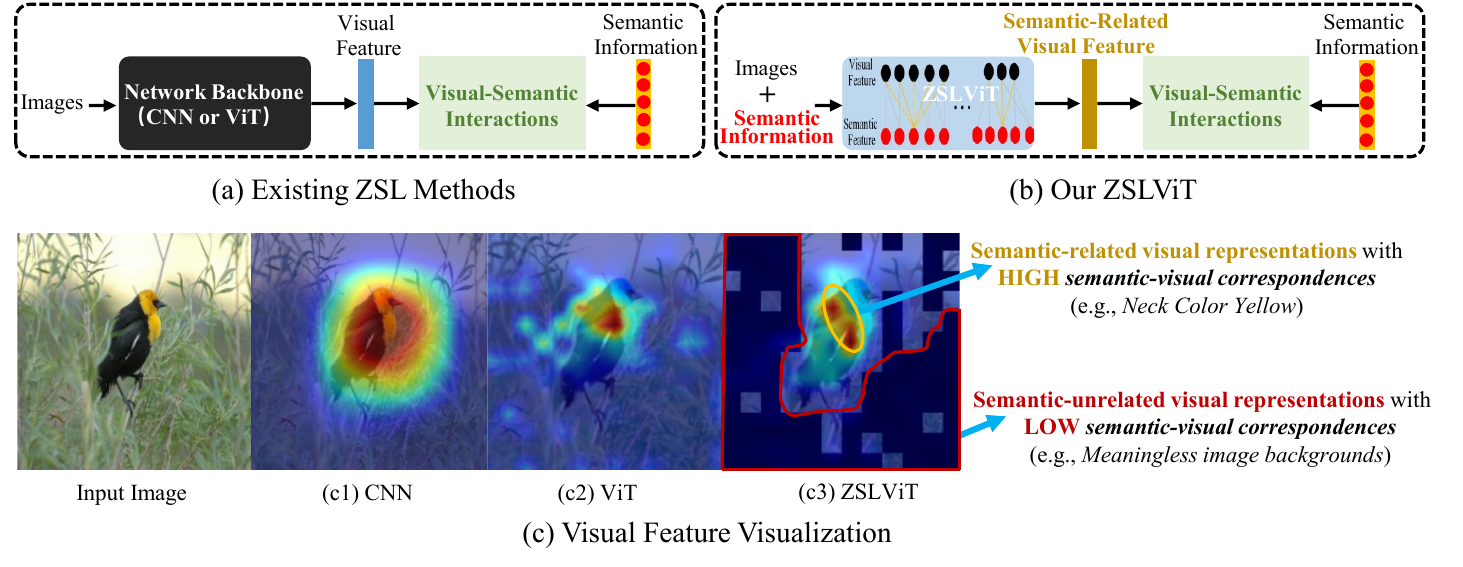}
			\caption{Motivation Illustration. (a) Existing ZSL methods simply take the pre-trained network backbone (\textit{i.e.}, CNN or ViT) to extract visual features. (b) Our ZSLViT progressively learns semantic-visual correspondences to represent semantic-related visual features in the \textit{whole network} for advancing ZSL. (c) The visual feature visualization. (\textit{c1}) The heat map of visual features learned by CNN backbone (\textit{e.g.}, ResNet101 \cite{He2016DeepRL}) includes the whole object and background, which fail to capture the semantic attributes. (\textit{c2}) The attention map of visual features learned by the standard ViT \cite{Dosovitskiy2020AnII}, which localizes the semantic attributes incorrectly. (\textit{c3}) The attention map learned by our ZSLViT, which discovers the semantic-related visual representations and discards the semantic-unrelated visual information according to semantic-visual correspondences.}
			\label{fig:motivation}
		\end{center}
	\end{figure*}

	Existing ZSL methods \cite{Xian2018FeatureGN,Xian2019FVAEGAND2AF,Wan2019TransductiveZL,Chen2021FREE, Chen2021HSVA,Chen2021TransZeroCA,Xie2019AttentiveRE, Alamri2021MultiHeadSV, Alamri2021ImplicitAE,Naeem2022I2DFormerLI,Liu2023ProgressiveSM,Kim2022SemanticFE,Chen2022DUETCS} typically take a network backbone (convolutional neural network (CNN) or vision Transformer (ViT)) pre-trained on ImageNet \cite{Russakovsky2015ImageNetLS} to extract visual features. However, the network backbone fails to learn matched visual-semantic correspondences for representing semantic-related visual features, because they lack sufficient guidance of semantic information, as shown in Fig. \ref{fig:motivation}(a).  As shown in Fig. \ref{fig:motivation}(c1), the CNN backbone learns the representations focused on the meaningless background information or the whole object. Although some methods adopt the attention mechanism to enhance the CNN visual features via attribute localization \cite{Xie2019AttentiveRE,Zhu2019SemanticGuidedML,Xu2020AttributePN,Liu2021GoalOrientedGE, Chen2021TransZeroCA,Chen2022MSDN,Narayan2021DiscriminativeRM}, they only obtain the sub-optimal visual representations as the visual spaces are almost fixed after the CNN backbone learning.
	
	Thanks to the strong capability of modeling long-range association of whole image, some methods simply take the pre-trained ViT to extract visual features for ZSL tasks \cite{Alamri2021MultiHeadSV, Alamri2021ImplicitAE,Naeem2022I2DFormerLI,Liu2023ProgressiveSM,Chen2022DUETCS} and achieve better performance than CNN features-based ZSL methods. Unfortunately, they localize the semantic attribute incorrectly without explicit guidance of semantic information, which also fails to represent the correspondences between visual-semantic features, as shown in Fig. \ref{fig:motivation}(c2). Therefore, the visual features learned by CNN or ViT backbone cannot be well related to their corresponding semantic attributes (\textit{e.g.}, the ‘\textit{neck color yellow}’ of Yellow\_Headed\_Blackbird), resulting in undesirable visual-semantic interactions. Consequently, the semantic knowledge transferring in ZSL is limited, thus leading to inferior ZSL performance. As such, properly constructing matched visual-semantic correspondences for learning semantic-related visual features in the feature extraction network for advancing ZSL is highly necessary.

	To learn semantic-related visual features for desirable visual-semantic interactions, we propose a \textit{progressive semantic-guided vision transformer} specifically for ZSL, dubbed ZSLViT. As shown in Fig. \ref{fig:motivation}(b) and (c3), ZSLViT takes two considerations in the whole network: i) how to discover the semantic-related visual representations explicitly, and ii) how to discard the semantic-unrelated visual information (\textit{e.g.}, meaningless image backgrounds). We first introduce a semantic-embedded token learning (SET) mechanism consisting of a semantic enhancement and a semantic-guided token attention. The semantic enhancement improves semantic-visual correspondences for visual tokens via visual-semantic consistency learning and semantic embedding. Accordingly, the semantic-guided token attention explicitly discovers the semantic-related visual tokens, which have high visual-semantic correspondences and are preserved into the next layer. Then, we introduce visual enhancement (ViE) to fuse the visual tokens with low visual-semantic correspondences into one new token for purifying the semantic-unrelated information. Thus, the semantic-unrelated visual information is discarded for enhancing visual features. These two operations are integrated into various encoders to progressively learn semantic-related visual representations, enabling desirable visual-semantic interactions for ZSL task. The quantitative and qualitative results demonstrate the superiority and great potential of ZSLViT.
	
	Our main contributions can be summarized:
	\begin{itemize}
		\item We propose a progressive semantic-guided visual transformer, dubbed ZSLViT, which learns matched visual-semantic correspondences for representing semantic-related visual representations, enabling effective visual-semantic interactions for ZSL.
		
		\item We introduce semantic-embedded token learning and visual enhancement to discover the semantic-related visual representations explicitly and discard the semantic-unrelated visual information, respectively.
		
		\item  We conduct extensive experiments on three challenging benchmark datasets (\textit{i.e.}, CUB \cite{Welinder2010CaltechUCSDB2}, SUN \cite{Patterson2012SUNAD}, and AWA2 \cite{Xian2019ZeroShotLC}) under both conventional and generalized ZSL settings. Results show that our  ZSLViT achieves significant improvements and new state-of-the-art results.
	\end{itemize}

	\section{Related Works}\label{Sec2}\vspace{-2mm}

	\noindent\textbf{Zero-Shot Learning.}
	ZSL typically transfers semantic knowledge from seen classes to unseen ones by conducting visual-semantic interactions, and thus the unseen classes can be recognized \cite{Akata2016LabelEmbeddingFI,Xian2018FeatureGN,Xian2019FVAEGAND2AF,Chen2021FREE, Han2021ContrastiveEF,Chen2023EGANSEG,Hassan2023AlignYP}. There two methods are typically adopted, \textit{i.e.}, embedding-based methods \cite{Liu2018GeneralizedZL,Liu2023ProgressiveSM,Chen2022DUETCS,Chen2021TransZero, Chen2021TransZeroCA,Chen2022MSDN, Huynh2020FineGrainedGZ} and generative methods \cite{Xian2019FVAEGAND2AF, Gupta2021GenerativeMZ,Hong2022SemanticCE, Chen2021HSVA, Huynh2020CompositionalZL, Narayan2020LatentEF, Chen2023EvolvingSP}. Embedding-based methods map visual features into semantic space and match them with their corresponding semantic prototypes by nearest-neighbor matching. The generative ZSL methods learn a generator conditioned by the semantic prototypes to synthesize the visual features for unseen classes, which are utilized to train a supervised classifier (\textit{e.g.}, softmax). Different from the generic image classification task that classifies classes based on the semantic-unrelated labels, ZSL aims to classify the unseen class samples according to the semantic prototypes that are represented by the specific semantic attributes. Thus, discovering semantic-related visual representations and discarding the semantic-unrelated visual information to conduct effective semantic knowledge transferring from seen classes to unseen ones for ZSL is very necessary. These methods take pre-trained CNN backbone (\textit{e.g.}, ResNet101) to extract the global visual features, which cannot accurately capture the semantic information of visual appearances (\textit{e.g.}, the ‘\textit{neck color yellow}’ of Yellow\_Headed\_Blackbird), resulting in undesirable visual-semantic interactions for semantic knowledge transferring. Thus their results are essentially limited.  
	
	Although some methods take attention mechanism \cite{Xie2019AttentiveRE,Zhu2019SemanticGuidedML,Xu2020AttributePN,Liu2021GoalOrientedGE, Chen2021TransZeroCA} to refine the extracted visual features from CNN backbone, they obtain the sub-optimal visual representations as the visual spaces are almost fixed after the CNN backbone learning. Considering that vision Transformer (ViT) \cite{Dosovitskiy2020AnII, Rao2021DynamicViTEV,Liang2022NotAP,Zeng2022NotAT} has the advantages of learning implicitly semantic-context visual information using self-attention mechanisms, in this work, we devise a novel ViT backbone to progressively learn the semantic-related visual features under the guidance of semantic information in the whole network. This encourages the model to conduct effective visual-semantic interactions in ZSL.

	\noindent\textbf{Vision Transformer.} 
	Transformers \cite{Vaswani2017AttentionIA} have achieved significant progress in computer vision recently due to their strong capability of modeling long-range relation, \textit{e.g.}, image classification \cite{Hu2019LocalRN}, object detection \cite{Carion2020EndtoEndOD}, and semantic segmentation \cite{Cheng2021PerPixelCI}. Vision Transformer (ViT) \cite{Dosovitskiy2020AnII} is the first pure Transformer backbone introduced for image classification, and it is further employed for other vision tasks \cite{Rao2021DynamicViTEV}. Some methods simply take ViT to extract the global visual features for ZSL tasks \cite{Alamri2021MultiHeadSV, Alamri2021ImplicitAE,Liu2023ProgressiveSM,Chen2022DUETCS}. Unfortunately, they fail to construct matched visual-semantic correspondences explicitly with semantic information and cannot well explore the potential of ViT for ZSL. In this work, we aim to design a ViT backbone specifically for advancing ZSL considering two properties: i) discover the semantic-related visual representations explicitly, and ii) discard the semantic-unrelated visual information. 
	\vspace{-2mm}
	
	\section{Semantic-Guided Vision Transformer}\label{Sec3}\vspace{-2mm}
	
	The task of ZSL is formulated as follows. Let we have $C^s$ seen classes data $\mathcal{D}^{s}=\left\{\left(x_{i}^{s}, y_{i}^{s}\right)\right\}$, where $x_i^s \in \mathcal{X}$ denotes the $i$-th sample, and $y_i^s \in \mathcal{Y}^s$ is its class label. The $\mathcal{D}^{s}$ is split into a training set $\mathcal{D}_{tr}^{s}$ and a testing set $\mathcal{D}_{te}^{s}$ following~\cite{Xian2019ZeroShotLC}. Meanwhile, we have $C^u$ unseen classes data $\mathcal{D}_{te}^{u}=\left\{\left(x_{i}^{u}, y_{i}^{u}\right)\right\}$, where $x_{i}^{u}\in \mathcal{X}$ is the sample of unseen classes, and $y_{i}^{u} \in \mathcal{Y}^u$ is its class label. Thus, the total class number in one dataset is $c \in \mathcal{C}^{s} \cup \mathcal{C}^{u}$. The semantic prototypes are represented by vectors. Each vector corresponds to one class.  Each semantic vector $z^{c}=\left[z^{c}(1), \ldots, z^{c}(A)\right]^{\top} \in \mathbb{R}^{|A|}$ is with the $|A|$ dimension, where each dimension is a semantic attribute value annotated by human. In the conventional ZSL setting (CZSL) setting, we learn a classifier only for unseen classes (\textit{i.e.}, $f_{\rm{CZSL}}: \mathcal{X} \rightarrow \mathcal{Y}^{U}$). Differently in generalized ZSL (GZSL), we learn a classifier for both seen and unseen classes (\textit{i.e.}, $f_{\rm{GZSL}}: \mathcal{X} \rightarrow \mathcal{Y}^{U} \cup \mathcal{Y}^{S}$).
	
\begin{figure*}[t]
		\begin{center}
			\includegraphics[width=16cm,height=5.8cm]{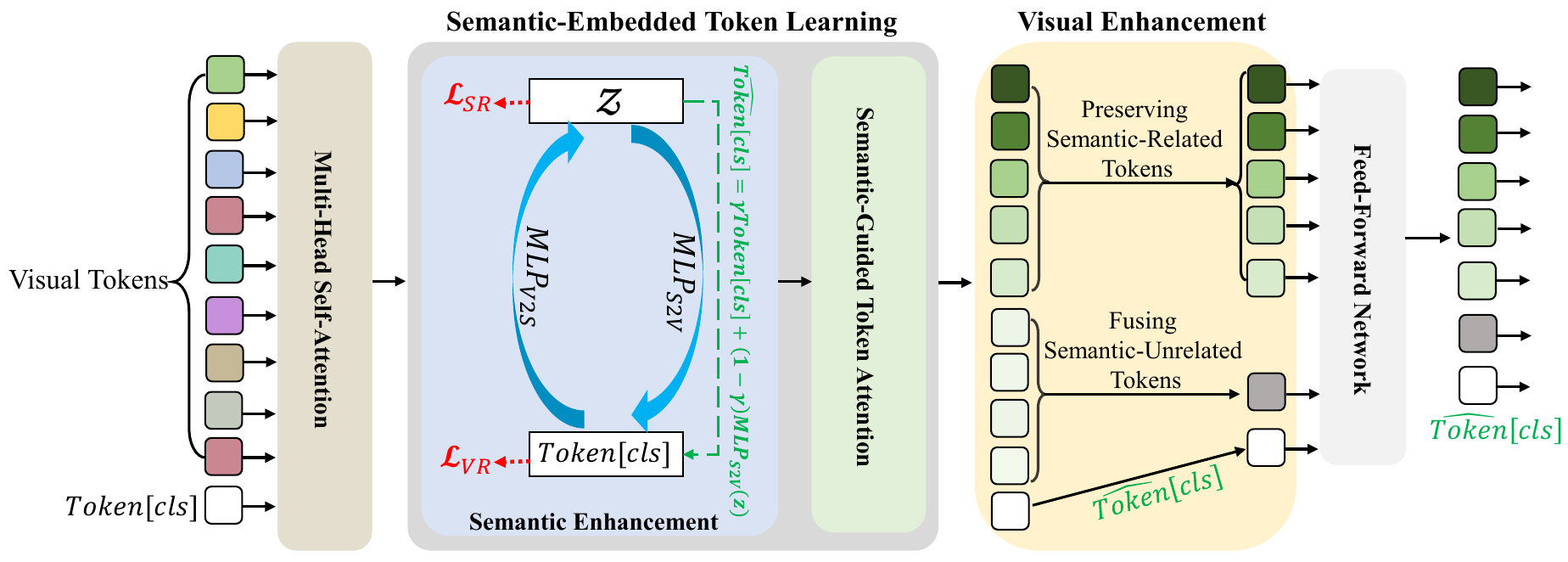}
			\caption{A single ZSLViT encoder. ZSLViT encoder includes a semantic-embedded token learning (SET) and a visual enhancement (ViE) between the multi-head self-attention and feed-forward network layers. SET improves the visual-semantic correspondences via semantic enhancement and discovers the semantic-related visual tokens explicitly with semantic-guided token attention. ViE fuses the visual tokens of low visual-semantic correspondences to discard the semantic-unrelated visual information for visual tokens enhancement. The ZSLViT encoder are integrated into various layers to progressively learn semantic-related visual representations, enabling effective visual-semantic interactions for ZSL.}
			\label{fig:framework}
		\end{center}
	\end{figure*}

	In the following, we introduce our ZSLViT specifically. As shown in Fig. \ref{fig:framework}, the novel operations of ZSLViT include a semantic-embedding token learning (SET) and a visual enhancement (ViE). These two operations are integrated into various encoders between the multi-head self-attention and feed-forward network layers to progressively learn semantic-related visual representations, enabling accurate visual-semantic interactions for ZSL. At the end of this section, we demonstrate how we perform zero-shot prediction using the semantic-related visual representations learned by ZSLViT.

	\subsection{Semantic-Embedded Token Learning}\vspace{-2mm}
	Semantic-embedded token learning (SET) is employed to discover semantic-related visual features. SET consists of a semantic enhancement module and a semantic-guided token attention module. The semantic enhancement explicitly improves the visual-semantic correspondences with visual-semantic consistency learning and semantic embedding. The semantic-guided token attention discovers the semantic-related visual representations based on the semantic-enhanced tokens.

	\noindent\textbf{Semantic Enhancement.} 
	We first conduct visual-semantic consistency learning based on the visual features and semantic vectors. Here, we take the $[cls]$ token (\textit{i.e.}, $token[cls]$ ) as the visual features due to it pays more attention (\textit{i.e.}, having a larger attention value) on class-specific tokens to represent one image for classification \cite{Vaswani2017AttentionIA,Caron2021EmergingPI}. Specifically, we take two multi-layer perceptrons (MLP), \textit{i.e.}, $MLP_{V2S}$ and $MLP_{S2V}$, to map the features from visual space to semantic space (\textit{i.e.}, $Visual \rightarrow Semantic$) and from semantic space to visual space  (\textit{i.e.}, $ Semantic\rightarrow Visual$), respectively. As such, the $MLP_{V2S}$ and $MLP_{S2V}$ can effectively improve their consistency. 
	\begin{align}
	\label{eq:consistency-learning}
	&Visual\rightarrow Semantic: \tilde{z} = MLP_{V2S}(Token[cls]), \\
	&Semantic\rightarrow Visual: \widetilde{Token}[cls] = MLP_{S2V}(z),
	\end{align}
	where $\tilde{z}$ is the reconstructed semantic vector from visual space, and $\widetilde{Token}[cls]$ is the reconstructed visual feature from semantic space. To enable visual-semantic consistency learning, we take a semantic reconstruction loss $\mathcal{L}_{SR}$ and a visual reconstruction loss $\mathcal{L}_{VR}$ to guide the optimization of $MLP_{V2S}$ and $MLP_{S2V}$, formulated as:
	\begin{align}
	\label{eq:sr}
	&\mathcal{L}_{SR} = \|z-\tilde{z}\|_1, \\
	\label{eq:vr}
	&\mathcal{L}_{VR} = \|Token[cls]-\widetilde{Token}[cls]\|_1.
	\end{align}
	Notably, we set a larger weight on $\mathcal{L}_{VR}$ than $\mathcal{L}_{SR}$ as we mainly aim to enhance semantic information into visual representations for subsequent learning. This will also facilitate stable optimization for $MLP_{V2S}$ and $MLP_{S2V}$.
	
	Considering the semantic vectors are informative attribute representations, we explicitly boost the semantic information into visual features for semantic enhancement via semantic embedding. Specifically, we concatenate the reconstructed visual features from semantic space with the real visual features $Token[cls]$:
	\begin{gather}
	\label{eq:semantic-embedding}
	\widehat{Token}[cls]= \gamma Token[cls] + (1-\gamma)MLP_{S2V}(z),
	\end{gather}
	where $\gamma$ is a combination coefficient, which is set to a relatively large value for progressive enhancement, enabling stable learning for ZSLViT. $\widehat{Token}[cls]$ is the semantically enhanced token, which is served as a new $Token[cls]$ (\textit{i.e.}, $Token[cls]=\widehat{Token}[cls]$) to update the original $Token[cls]$ for subsequent learning. We should note that semantic embedding is only used in the training stage but not in the inference stage. 
	
	\noindent\textbf{Semantic-Guided Token Attention.} 
	After semantic enhancement, we take semantic-guided token attention to identify the semantic-related and semantic-unrelated visual tokens based on $\widehat{Token}[cls]$. Specifically, we perform the interaction between the $\widehat{Token}[cls]$ and other visual tokens, where the packed outputs of the multi-head self-attention layer are used as keys ($K$) and values ($V$), and $\widehat{Token}[cls]$ is the query vector. It is defined as:
	\begin{gather}
	\label{eq:semantic-attention}
	f(x)=\operatorname{Softmax}\left(\frac{\widehat{Token}[cls] \cdot K^{\top}}{\sqrt{d}}\right) V=a \cdot V,
	\end{gather}
	where $d$ is a scaling factor. $a=\{a_1,a_2,\cdots, a_n\}$ ($n$ is the number of input visual tokens in a ZSLViT encoder) is \textit{attention scores being visual-semantic correspondences} from $[cls]$ token to other visual tokens. Accordingly,  $f(x)$ is a linear combination of the value vectors $V=\{v_1,v_2,\cdots, v_n\}$. Since $v_i$ comes from the $i$-th visual token, the attention score $a_i$ determines how much information of the $i$-th visual token is embedded into the output of $[cls]$ token. It is natural to assume that the visual-semantic correspondence $a_i$ indicates the importance of the $i$-th token corresponding to the semantic information for visual representations. To this end, ZSLViT can effectively localize the image regions most relevant to semantic attributes for discovering the semantic-related visual features, as shown in Fig. \ref{fig:motivation}(c3).

	\subsection{Visual Enhancement}
	Visual enhancement (ViE) is devised to discard the semantic-unrelated visual features for enhancing visual features further. According to the visual-semantic correspondences $a$ in Eq. \ref{eq:semantic-attention}, ZSLViT can easily determine the semantic-related visual tokens (\textit{i.e.}, with the Top-$k$ largest $a$, and the indices set denoted as $\mathcal{P}$), and the semantic-unrelated visual tokens(\textit{i.e.}, with the $n-k$ smallest $a$, and the indices set denoted as $\mathcal{N}$). We take a hyper-parameter $\kappa=k/n$ to determine $\mathcal{P}$ and $\mathcal{N}$. Since visual-semantic interactions in ZSL rely on semantic-related visual information, we can preserve the semantic-related visual tokens and discard the semantic-unrelated visual tokens to alleviate the negative effects of meaningless visual information (\textit{e.g.}, the background of image). Thus the visual features are enhanced to enable effective visual-semantic interactions in ZSL. Considering ZSLViT cannot completely learn the accurate semantic-related visual representation in an encoder at one time, we fuse the semantic-unrelated visual tokens at the current stage to supplement semantic-related ones:
	\begin{gather}
	\label{eq:fusing-features}
	T(x)=\{f(x)_i\}_{i=1}^{\mathcal{P}} \cup \sum_{j \in \mathcal{N}} a_j f(x)_j,
	\end{gather} 
	$T(x)$ is the semantic-related visual features in the current encoder and is used for subsequent learning in the feed-forward network layer and next encoder. Thus, ZSLViT purifies the visual tokens in various encoders to discard the meaningless visual information progressively. Meanwhile, ZSLViT can be effectively lightened to reduce computational costs, enabling model acceleration.
	
	\subsection{Model Optimization}
	We now introduce the optimization objectives of our ZSLViT. First, ZSLViT conducts semantic-embedded token learning in various layers/encoders (indexed by $S$), where include $\mathcal{L}_{SR}$ (Eq. \ref{eq:sr}) and $\mathcal{L}_{SR}$ (Eq. \ref{eq:vr}). Assuming we are dealing with a minibatch of $B$
	samples $x_i\in \mathcal{D}_{tr}^s$, it can be formulated as:
	\begin{gather}
	\label{eq:sr-vr}
	\mathcal{L}_{SET}=\frac{1}{B}\frac{1}{S} \sum_{i=1}^{B} \sum_{s=1}^{S}(\lambda_{SR}\mathcal{L}_{SR}^s(x_i)+\lambda_{VR}\mathcal{L}_{VR}^s(x_i)).
	\end{gather}

	Further, ZSLViT also trains the prediction module at the last layer such that it can produce favorable predictions and fine-tune the backbone to make it adapt to semantic-related visual feature learning.
	Since our ZSLViT is an embedding-based model, we should map the visual features (\textit{i.e.}, the $token[cls]$ in the last layer) into their corresponding semantic vectors, \textit{i.e.}, $\phi(x_i)= Token[cls]^{\top} {W}_{V2S}$, where ${W}_{V2S}$ is a learnable mapping matrix. Following \cite{Chen2022MSDN}, we also employ the attribute-based cross-entropy loss  for optimization:
	\begin{gather}
	\label{eq:L_ACEC}\
	\mathcal{L}_{pre}=-\frac{1}{B} \sum_{i=1}^{B} \log \frac{\exp \left(\phi(x_i) \times z^{c}\right)}{\sum_{\hat{c} \in \mathcal{C}^s} \exp \left(\phi(x_i)\times z^{\hat{c}} \right)}.
	\end{gather}
	Different from existing embedding-based ZSL methods \cite{Xu2020AttributePN, Chen2021TransZero,Chen2022MSDN,Alamri2021ImplicitAE,Alamri2021MultiHeadSV} that take an additional self-calibration loss to tackle the seen-unseen bias problem \cite{Chen2021FREE} during training, our ZSLViT can automatically avoid this issue as the discovered semantic-related visual features have good generalization from seen classes to unseen ones.

	To this end, the overall loss function of ZSLViT is defined as:
	\begin{gather}
	\label{eq:L_all}
	\mathcal{L}_{\textit{ZSLViT}} =\mathcal{L}_{SET} + \mathcal{L}_{pre}.
	\end{gather}
	As the $\mathcal{L}_{pre}$ is the base loss in embedding-based ZSL, we set its weight to one.
	
	\subsection{Zero-Shot Prediction}
	We conduct zero-shot prediction in the inference stage. We first obtain the embedding features $\phi(x_i)$ of a test
	instance $x_i$ from testing set (\textit{i.e.}, $x_i\in \mathcal{D}_{te}^s\cup \mathcal{D}_{te}^u$) in the semantic space. Then, we take an explicit calibration to predict the test label of $x_i$, which is formulated as:
	\begin{gather}
	\label{Eq:prediction}
	c^{*}=\arg \max _{c \in \mathcal{C}^u/\mathcal{C}}softmax(\phi(x_i) \times z^{c})+\mathbb {I}_{\left[c\in\mathcal{C}^u\right]}.
	\end{gather}
	$\mathbb {I}_{\left[c\in \mathcal{C}^u\right]}$ is an indicator function (\textit{i.e.}, it is $\tau$ when $c\in\mathcal{C}^u$, otherwise zero, where $\tau$ is a hyper-parameter to control the calibration degree). We empirically set $\tau$ to 0.4 for all datasets. $\mathcal{C}^u/\mathcal{C}$ corresponds to the CZSL/GZSL setting, respectively. Since we do not use the unlabeled samples of unseen classes during training, our ZSLViT is an inductive method.

	\begin{table*}[t]
		\centering  
		\caption{State-of-the-art comparisons on CUB, SUN, and AWA2 under the CZSL and GZSL settings. CNN features-based ZSL methods are categorized as $\sharp$, and ViT features-based ZSL methods are categorized as $\flat$. The symbol “*” denotes attention-based ZSL methods using CNN features. The symbol $\dagger$ denotes ZSL methods based on large-scale vision-language model. The best and second-best results are marked in \textbf{\color{red}Red} and \textbf{\color{blue}Blue}, respectively.}
		\resizebox{\linewidth}{!}{\small
			\begin{tabular}{r|r|c|c|ccl|c|ccl|c|ccl}
				\hline
				&\multirow{3}{*}{\textbf{Methods}}& \multirow{3}{*}{\textbf{Venue}}
				&\multicolumn{4}{c|}{\textbf{CUB}}&\multicolumn{4}{c|}{\textbf{SUN}}&\multicolumn{4}{c}{\textbf{AWA2}}\\
				\cline{4-7}\cline{8-11}\cline{12-15}
				&&&\multicolumn{1}{c|}{CZSL}&\multicolumn{3}{c|}{GZSL}&\multicolumn{1}{c|}{CZSL}&\multicolumn{3}{c|}{GZSL}&\multicolumn{1}{c|}{CZSL}&\multicolumn{3}{c}{GZSL}\\
				\cline{4-7}\cline{8-11}\cline{12-15}
				&&&\rm{acc}&\rm{U} & \rm{S} & \rm{H} &\rm{acc}&\rm{U} & \rm{S} & \rm{H} &\rm{acc}&\rm{U} & \rm{S} & \rm{H} \\
				\hline
				\multirow{2}*{
					\begin{tabular}{c}
						\vspace{-5cm}$\sharp$
					\end{tabular}
				}
				&AREN$^{*}$~\cite{Xie2019AttentiveRE}&CVPR'19&71.8&38.9&\textbf{\color{red}78.7}&52.1&60.6&19.0&38.8&25.5&67.9&15.6&\textbf{\color{blue}92.9}&26.7 \\
				&f-VAEGAN~\cite{Xian2019FVAEGAND2AF}&CVPR'19&61.0&48.4&60.1& 53.6&64.7&45.1&38.0&41.3&71.1&57.6&70.6&63.5\\
				&TF-VAEGAN~\cite{Narayan2020LatentEF}&ECCV'20&64.9&52.8&64.7& 58.1&\textbf{\color{blue}66.0}&45.6&40.7& 43.0&72.2&59.8&75.1& 66.6\\
				&LsrGAN~\cite{Vyas2020LeveragingSA}&ECCV'20&--&48.1&59.1& 53.0&--&44.8&37.7& 40.9&--&54.6&74.6& 63.0\\
				&DAZLE$^{*}$~\cite{Huynh2020FineGrainedGZ}&CVPR'20&66.0&56.7&59.6&58.1&59.4&52.3&24.3&33.2&67.9&60.3&75.7&67.1\\
				&APN$^{*}$~\cite{Xu2020AttributePN}&NeurIPS'20&72.0&65.3& 69.3&67.2&61.6& 41.9&34.0&37.6&68.4&57.1&72.4&63.9\\
				&Composer$^{*}$~\cite{Huynh2020CompositionalZL}&NeurIPS'20&69.4&56.4&63.8&59.9&62.6& \textbf{\color{red}55.1}&22.0& 31.4&71.5&62.1&77.3&68.8\\
				&FREE~\cite{Chen2021FREE}&ICCV'21&--&55.7&59.9&57.7&--& 47.4&37.2& 41.7&--& 60.4&75.4&67.1\\
				&GCM-CF~\cite{Yue2021CounterfactualZA}&CVPR'21&--&61.0&59.7&60.3&--& 47.9&37.8& 42.2&--& 60.4&75.1&67.0\\
				&HSVA~\cite{Chen2021HSVA}&NeurIPS'21&62.8&52.7&58.3&55.3&63.8&48.6&39.0&43.3&--&59.3&76.6&66.8\\
				&MSDN$^{*}$~\cite{Chen2022MSDN}&CVPR'22&76.1&68.7&67.5&68.1&65.8&52.2&34.2&41.3&70.1&62.0&74.5&67.7\\
				&GEM-ZSL$^{*}$~\cite{Liu2021GoalOrientedGE}&CVPR'22&\textbf{\color{blue}77.8}&64.8&77.1&\textbf{\color{blue}70.4}&62.8& 38.1&35.7&36.9&67.3&64.8&77.5&70.6\\
				&SE-GZSL~\cite{Kim2022SemanticFE}&AAAI'22 &--&53.1 &60.3& 56.4&--&45.8&40.7&  43.1&--& 59.9&80.7& 68.8\\ 
				&TransZero$^{*}$~\cite{Chen2021TransZero}&AAAI'22  &76.8&\textbf{\color{blue}69.3}&68.3&68.8&65.6&52.6&33.4&40.8&70.1&61.3&82.3&70.2\\
				&VS-Boost\cite{Li2023VSBoostBV}&IJCAI'23&--&68.0& 68.7& 68.4&--&49.2&37.4& 42.5&--&--&--&--\\
				&ICIS\cite{Christensen2023ImagefreeCI}&ICCV'23&60.6& 45.8& 73.7& 56.5&51.8&45.2& 25.6& 32.7&64.6& 35.6 &\textbf{\color{red}93.3}& 51.6\\

				\hline
				\multirow{2}*{
					\begin{tabular}{c}
						\vspace{-1cm}$\flat$
					\end{tabular}
				}
				&CLIP$^{\dagger}$~\cite{Radford2021LearningTV}&ICML'21&--& 55.2&54.8 &55.0&-- &--&--&--&-- &--&--&--\\
				&CoOp$^{\dagger}$~\cite{Zhou2022LearningTP}&IJCV'22&--& 49.2&63.8& 55.6&-- &--&--&--&-- &--&--&--\\
				&I2DFormer-Wiki~\cite{Naeem2022I2DFormerLI}&NeurIPS'22&45.4& 35.3& 57.6 &43.8&-- &--&--&--& \textbf{\color{red}76.4}&\textbf{\color{red}66.8}& 76.8& 71.5\\
				&CoOp+SHIP$^{\dagger}$\cite{Wang2023ImprovingZG}&ICCV'23& --& 55.3& 58.9& 57.1&-- &--&--&--&-- &--&--&--\\
				&I2MVFormer-Wiki~\cite{Naeem2022I2MVFormerLL}&CVPR'23&42.1& 32.4 &63.1&42.8&-- &--&--&--& \textbf{\color{blue}73.6}& \textbf{\color{blue}66.6}&82.9& \textbf{\color{blue}73.8}\\
				&DUET~\cite{Chen2022DUETCS}&AAAI'23&72.3& 62.9& 72.8& 67.5& 64.4& 45.7& \textbf{\color{blue}45.8}& \textbf{\color{blue}45.8}& 69.9& 63.7& 84.7&72.7\\
				\cline{2-15}
				&\textbf{ZSLViT} (\textbf{Ours})&--&\textbf{\color{red}78.9}&\textbf{\color{red}69.4}&\textbf{\color{blue}78.2}&\textbf{\color{red}73.6}	&\textbf{\color{red}68.3}&\textbf{\color{blue}45.9}&\textbf{\color{red}48.4}&\textbf{\color{red}47.3}&70.7&66.1&84.6&\textbf{\color{red}74.2} \\
				\hline
		\end{tabular} }
		\label{table:SOTA}
	\end{table*}
	
	\section{Experiments}\label{Sec4}\vspace{-2mm}

	\noindent\textbf{Benchmark Datasets.} 	
	We conduct extensive experiments on three popular ZSL benchmarks, including two fine-grained datasets (e.g., CUB \cite{Welinder2010CaltechUCSDB2} and SUN \cite{Patterson2012SUNAD}) and a coarse-grained dataset (e.g., AWA2 \cite{Xian2019ZeroShotLC}), to verify our ZSLViT. In specific, CUB includes 11,788 images of 200 bird classes (seen/unseen classes = 150/50) captured with 312 attributes. SUN consists of 14,340 images from 717 scene classes (seen/unseen classes = 645/72) captured with 102 attributes. AWA2 has 37,322 images from 50 animal classes (seen/unseen classes = 40/10) captured with 85 attributes.
	
	\noindent\textbf{Evaluation Protocols.}
	Following \cite{Xian2019ZeroShotLC}, we measure the top-1 accuracy both in the CZSL and GZSL settings. In the CZSL setting, we only measure the accuracy of the test samples from the unseen classes, \textit{i.e.}, $\bm{acc}$. In the GZSL setting, we compute the accuracy of the test samples from both the seen classes (denoted as $\bm{S}$) and unseen classes (denoted as $\bm{U}$). To generally evaluate the performance, the harmonic mean between seen and unseen classes is a main evaluation protocol in the GZSL setting, defined as $\bm{H =(2 \times S \times U) /(S+U)}$.
	
	\noindent\textbf{Implementation Details.}
	We use the training splits proposed in \cite{Xian2018FeatureGN}. For a fair comparison, we take the ViT-base model \cite{Touvron2021TrainingDI} pre-trained on ImageNet-1k as a baseline and for initialization.  The $MLP_{S2V}$ and $MLP_{V2S}$ has multiple hidden layer with ReLU activation. We incorporate our semantic-embedded token learning and visual enhancement operations into the 4-th, 7-th and 10-th encoder by default for CUB and AWA2 (4-th and 7-th layers for SUN) to progressively learn the semantic-related visual features. We use the Adam optimizer with hyper-parameters ($\beta_1$ = 0.9, $\beta_2$ = 0.999) to optimize our model. The batch size is set to 32. We empirically set $\gamma$ and $\kappa$ to 0.9 for all datasets.  We perform experiments on a single NVIDIA Tesla V100 graphic card with 32GB memory. We use PyTorch\footnote{\url{https://pytorch.org/}} for the implementation of all experiments.
	
	\begin{figure*}[t]
		\begin{center}
			\includegraphics[width=1.0\linewidth]{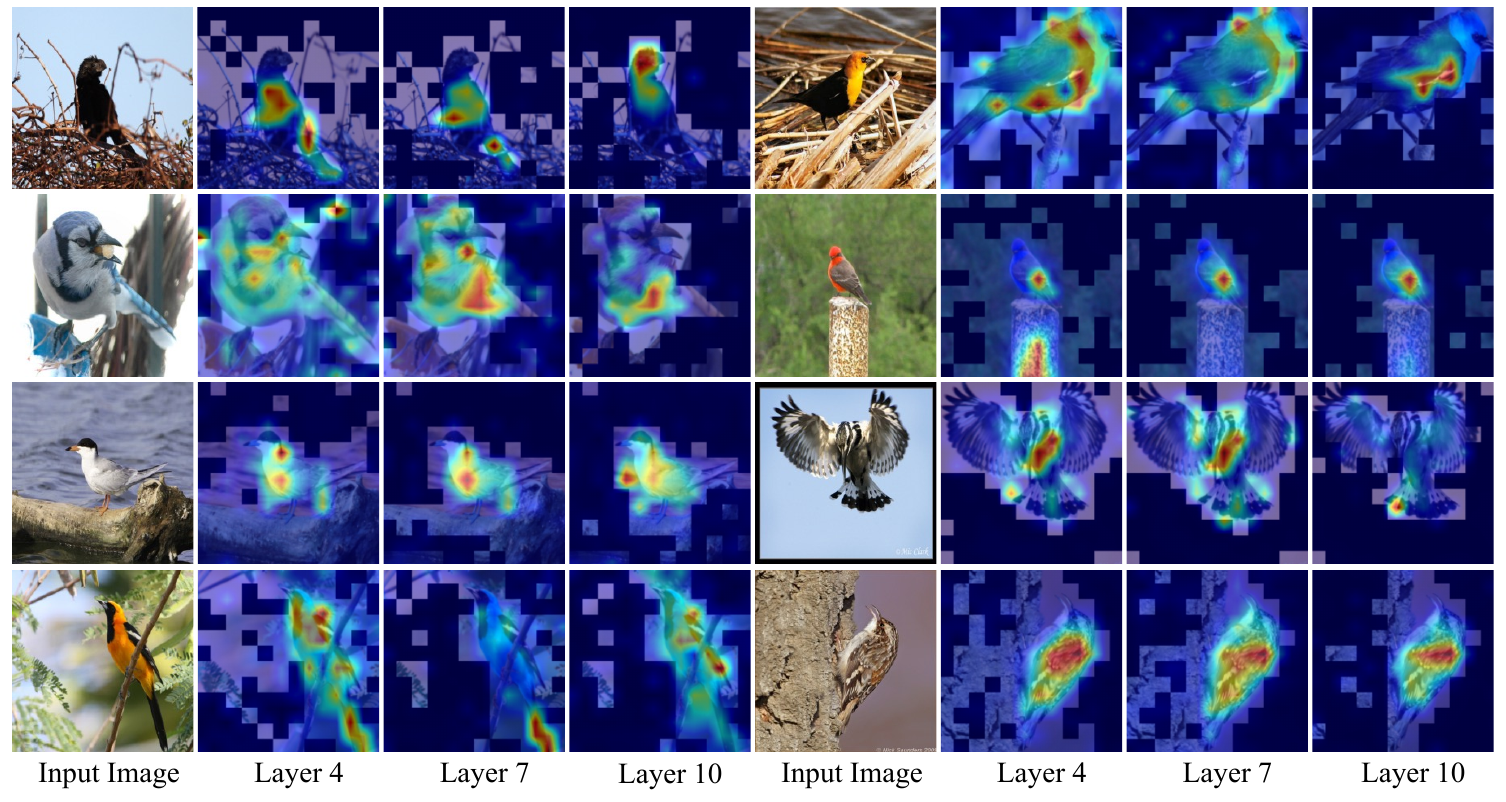}\\\vspace{-2mm}
			\caption{Visualizations of attention mask and map of our ZSLViT in various layers. The masked regions represent the semantic-unrelated visual tokens with low visual-semantic correspondences, which are fused into a new token for subsequent learning. The highlighted attention maps are the semantic-related visual tokens with high visual-semantic correspondences, which are preserved to next layer. Results show that ZSLViT can accurately identify the semantic-relateds/unrelated visual tokens in images for visual enhancement.}
			\label{fig:attention-mask}
		\end{center}
	\end{figure*}

	\subsection{Experimental Results}
	\noindent\textbf{Results of Conventional Zero-Shot Learning.} 
	Table \ref{table:SOTA}  shows the results of various methods on various datasets in the CZSL setting. Results show that the ViT features-based ZSL methods obtain better performances on all datasets than the CNN features-based ZSL methods. This is because ViT has obvious advantages of learning implicitly semantic-context visual information using a self-attention mechanism, which is well beneficial for ZSL task. Compared to all CNN features-based and ViT features-based methods, our ZSLViT consistently achieves the best performance of 78.9\% and 68.3\% on CUB and SUN datasets, respectively. Our ZSLViT obtains significant performance gains than the other ViT features-based methods \cite{Alamri2021ImplicitAE,Naeem2022I2DFormerLI,Chen2022DUETCS,Naeem2022I2MVFormerLL} on fine-grained datasets, \textit{i.e.}, by 6.6\% and 3.9\% on CUB and SUN respectively. Our significant performance improvements demonstrate that ZSLViT effectively learns the semantic-related visual features for desirable semantic knowledge transferring from seen classes to unseen ones. Since the coarse-grained dataset (\textit{e.g.}, AWA2) has a number of visual-unrelated semantic attributes for class descriptions (\textit{e.g.}, “\textit{eat fish}”), ZSLViT cannot achieve similar improvements on the AWA2.

	\noindent\textbf{Results of Generalized Zero-Shot Learning.} 
	Table \ref{table:SOTA} presents the GZSL performances of various methods, including CNN features-based and ViT features-based methods. Similar to the CZSL setting, ViT features-based ZSL methods perform better than CNN features-based methods generally on all datasets in the GZSL setting. Although some CNN features-based methods \cite{Zhu2019SemanticGuidedML,Xie2019AttentiveRE,Huynh2020FineGrainedGZ,Xu2020AttributePN,Huynh2020CompositionalZL,Chen2022MSDN,Liu2021GoalOrientedGE, Chen2021TransZero} take attention mechanism to localize semantic attribute for enhancing CNN features further, the visual space is relatively fixed after CNN backbone training, resulting in sub-optimal performance. This indicates the advantages of ViT in ZSL task further. Compared to all various methods, our ZSLViT obtains the best results on all datasets, \textit{e.g.}, 73.6\%, 47.3\% and 74.2\% on CUB, SUN and AWA2, respectively. Especially, our ZSLViT outperforms the latest ViT features-based ZSL method (\textit{i.e.}, I2MVFormer \cite{Naeem2022I2MVFormerLL}) by a large margin, resulting in harmonic mean improvements by 30.8\%. Notably, our ZSLViT also significantly outperforms the large-scale vision-language based ZSL methods (\textit{e.g.}, CLIP \cite{Radford2021LearningTV} and CoOP \cite{Zhou2022LearningTP}).  These results demonstrate that our ZSLViT can sufficiently explore the potential of ViT to learn semantic-related visual features for ZSL. 
	
	\begin{table}[t]
		\centering
		\caption{ Ablation studies for different components of ZSLViT on the CUB and AWA2 datasets.} \label{table:ablation}
		\resizebox{0.49\textwidth}{!}
		{
			\begin{tabular}{l|cc|cc}
				\hline
				\multirow{2}*{Method} &\multicolumn{2}{c|}{CUB} &\multicolumn{2}{c}{AWA2}\\
				\cline{2-3}\cline{4-5}
				&\rm{acc}& \rm{H} &\rm{acc}& \rm{H}\\
				\hline
				ZSLViT w/o SET ($\mathcal{L}_{VR}$)   &76.4&63.7
				&61.4&66.3\\
				ZSLViT w/o SET ($\mathcal{L}_{SR}$)  &78.0&72.8&70.2&73.6\\
				ZSLViT w/o SET (Eq. \ref{eq:semantic-embedding}) &76.2&69.7&69.9&71.2\\
				ZSLViT w/o ViE    & 77.9&72.4&70.5&73.9\\	
				ZSLViT (full) & \textbf{78.9}&\textbf{73.6}&\textbf{70.7}&\textbf{74.2}\\	
				\hline
			\end{tabular}
		}
	\end{table}
	
	\noindent\textbf{Ablation Study.} 
	We conduct ablation studies to evaluate the effectiveness of our ZSLViT in terms of the visual regression loss in SET (denoted as SET ($\mathcal{L}_{VR}$)), semantic regression loss in SET (denoted as SET ($\mathcal{L}_{SR}$)), semantic embedding in SET (denoted as SET ( Eq. \ref{eq:semantic-embedding})), and ViE. Results are shown in Table \ref{table:ablation}.  ZSLViT performs poorer slightly than its full model when no semantic reconstruction loss is used in SET, \textit{i.e.}, the acc/harmonic mean drop by 0.9\%/0.8\% on CUB and 0.5\%/0.6\% on AWA2. If visual reconstruction loss is not used in SET, ZSLViT achieves very poor results compared to its full model, \textit{i.e.}, the acc/harmonic mean drops by more than 2.5\%/9.9\% on CUB and 9.3\%/7.9\% on AWA2. These results show that visual reconstruction loss $\mathcal{L}_{VR}$ is essential for semantic-embedded token learning, because $\mathcal{L}_{VR}$ encourages SET to incorporate semantic information into visual tokens. This is typically neglected by existing ViT features-based ZSL methods \cite{Alamri2021ImplicitAE,Naeem2022I2DFormerLI,Chen2022DUETCS,Naeem2022I2MVFormerLL}. Moreover, ZSLViT cooperates with semantic embedding to improve its performance of acc/harmonic mean by 2.7\%/3.9\% on CUB and 0.8\%/3.0\% on AWA2, respectively. ViE further improves the performance of ZSLViT by discarding the semantic-unrelated visual tokens. Generally, these results indicate the effects of various components of ZSLViT.
	
	\begin{figure*}[t]
		\begin{center}
			\includegraphics[width=1.0\linewidth]{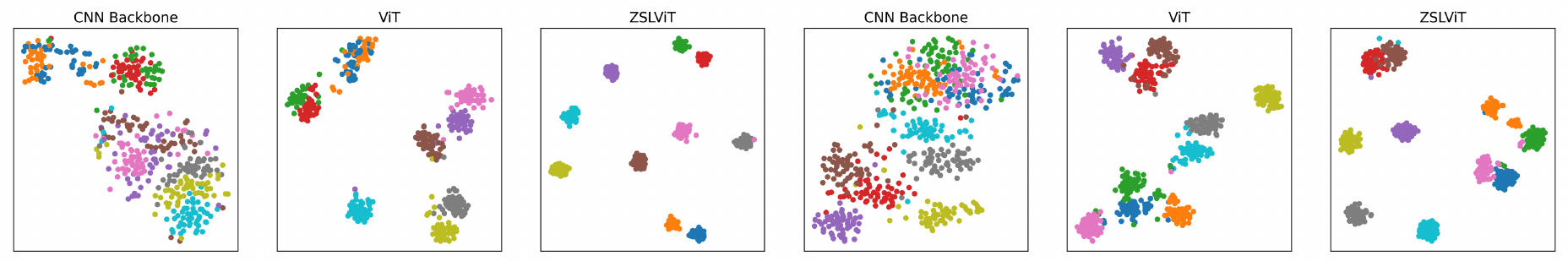}\\ 
			(a) Seen Classes \hspace{4cm} (b) Unseen Classes \\
			\caption{t-SNE visualizations of visual features for (a) seen classes and (b) unseen classes, learned by the CNN backbone (\textit{e.g.}, ResNet101 \cite{He2016DeepRL}), standard ViT \cite{Touvron2021TrainingDI} and our ZSLViT. The 10 colors denote 10 different seen/unseen classes randomly selected from CUB. (Best viewed in color)}
			\label{fig:tsne}
		\end{center}
	\end{figure*}
	
	\renewcommand{\tabcolsep}{1pt}
	\begin{figure*}[t]
		\begin{center}
			\begin{tabular}{cccc}
				\hspace{-5mm}\includegraphics[width=0.26\linewidth]{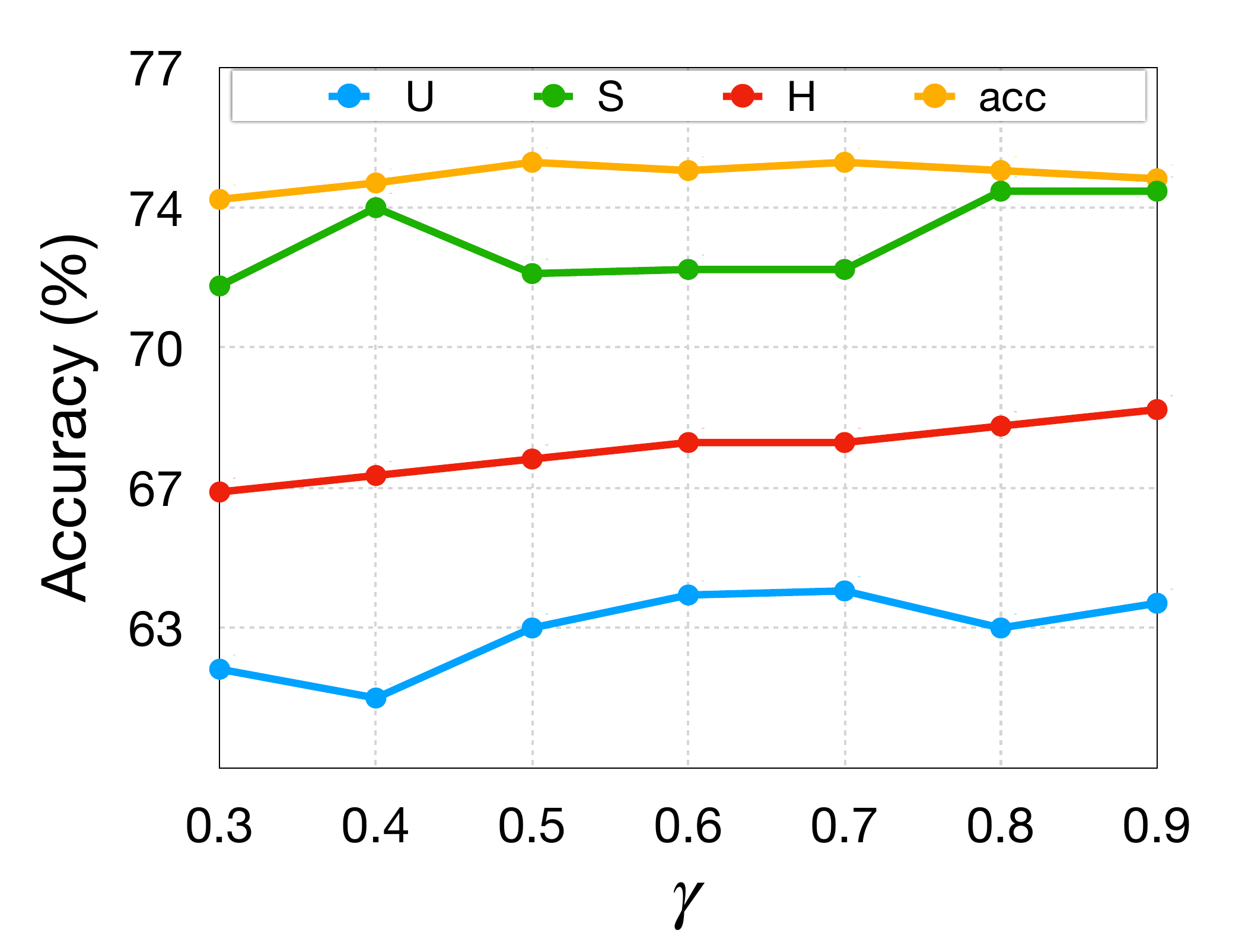}&
				\includegraphics[width=0.26\linewidth]{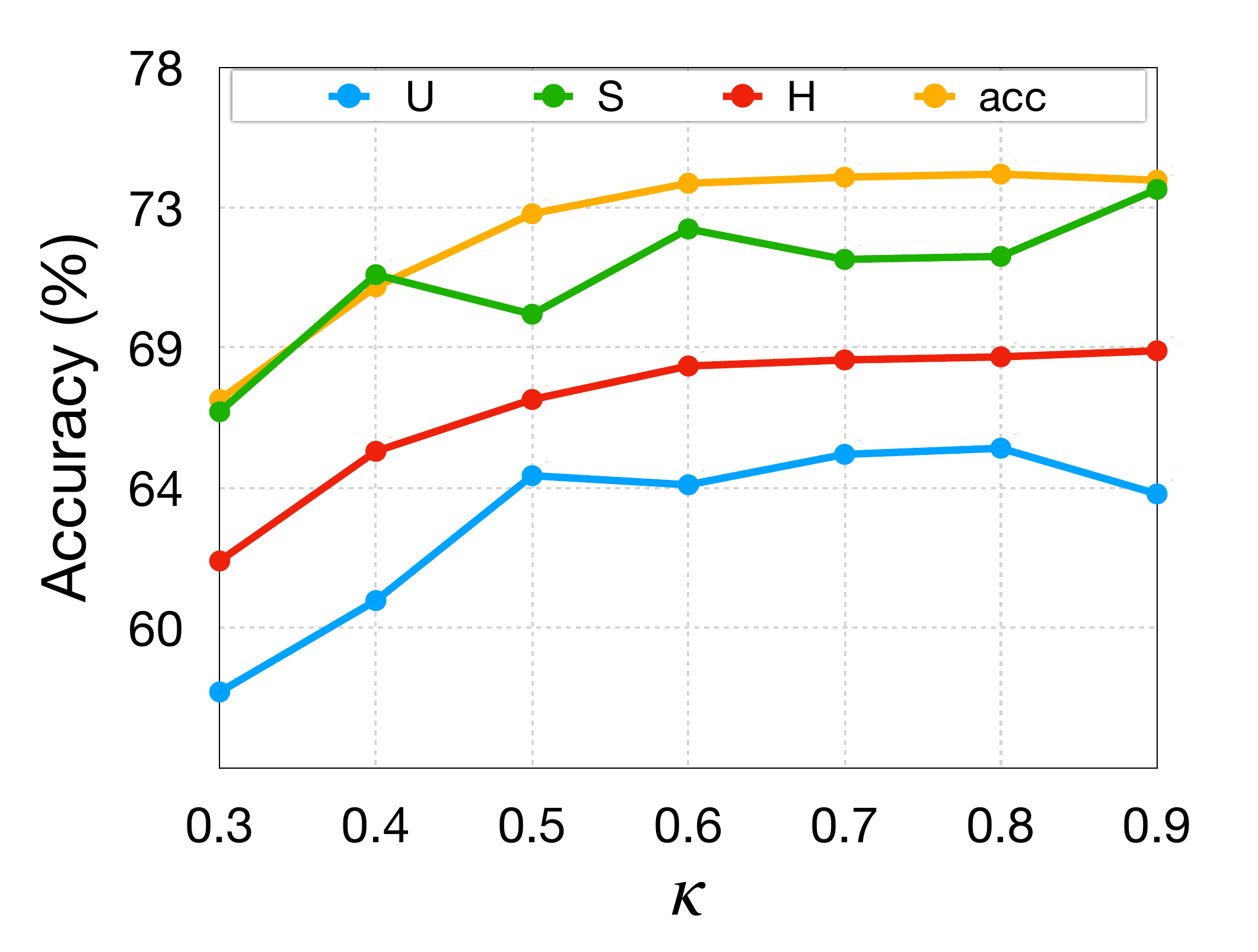}&
				\includegraphics[width=0.26\linewidth]{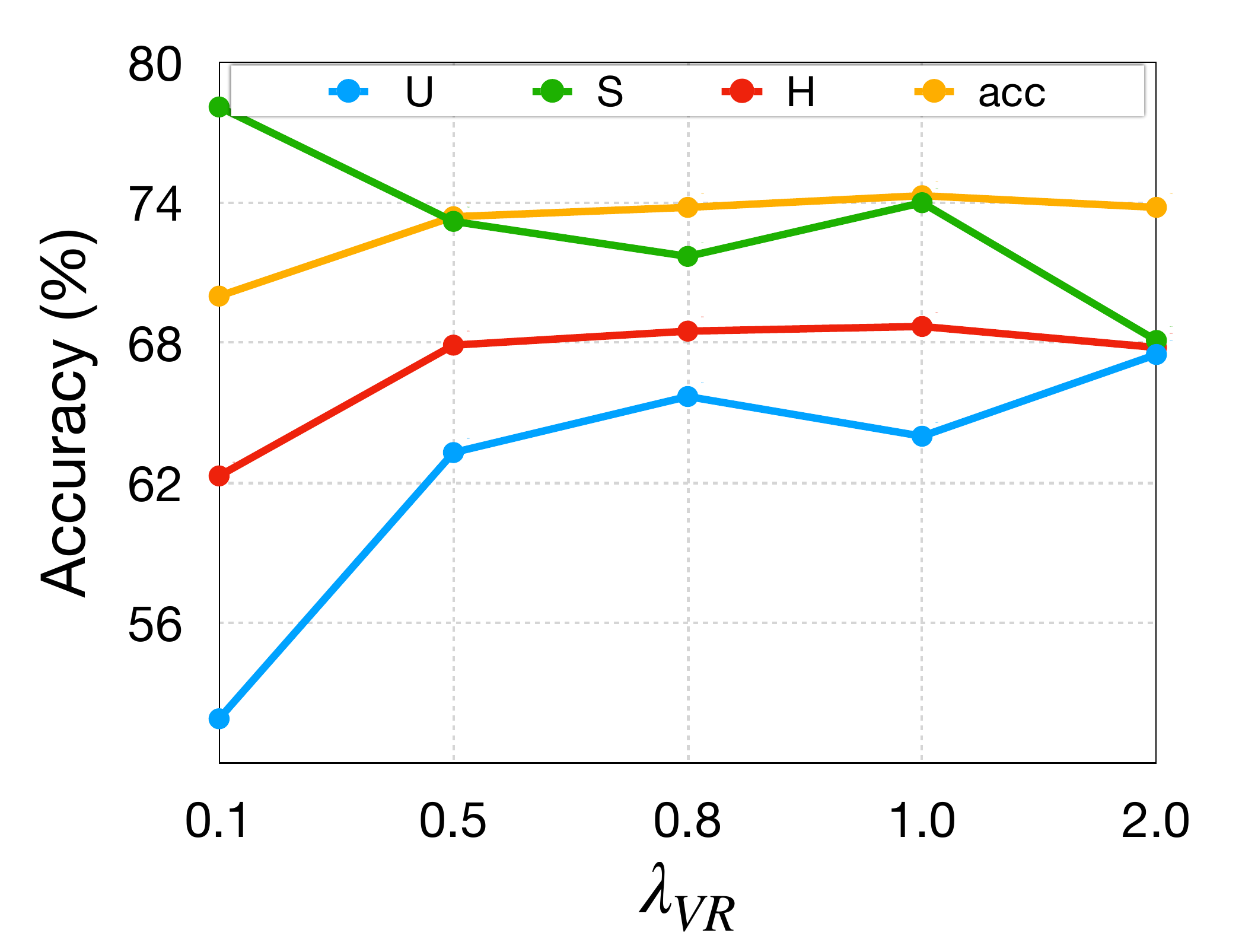}&
				\includegraphics[width=0.26\linewidth]{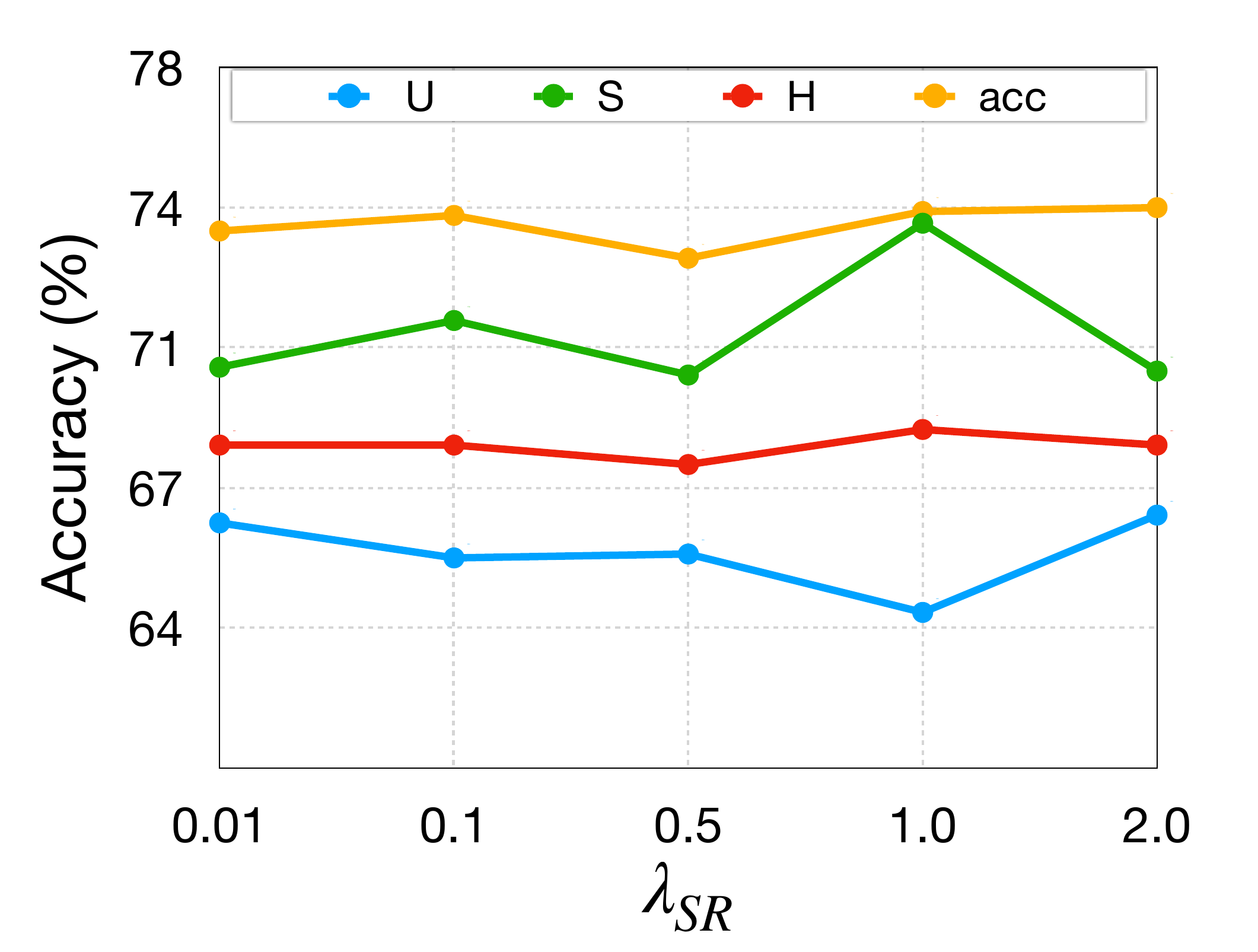}\\
				(a) Varying $\gamma$ effect& (b) Varying $\kappa$ effect& 
				(c) Varying $\lambda_{VR}$ effect& (d) Varying $\lambda_{SR}$ effect\\ 
			\end{tabular}
			\caption{The effects of (a) embedding coefficient $\gamma$, (b) fusing rate $\kappa$, (c) loss weights $\lambda_{VR}$, and (d) loss weights $\lambda_{SR}$. We take CUB as an example.}\label{fig:hyper-para}
		\end{center}
	\end{figure*}

	\noindent\textbf{Attention Mask and Map Visualization.}
	To intuitively show the effectiveness of our ZSLViT at learning semantic-related visual features for advancing ZSL, we visualize the attention mask (the tokens with low visual-semantic correspondences) and attention maps learned by ZSLViT on CUB. Results are shown in Fig. \ref{fig:attention-mask}. Thanks to the semantic-embedded token learning that identifies the semantic-related/unrelated tokens according to the semantic-guided token attention, ZSLViT effectively i) preserves the semantic-related visual tokens of high visual-semantic correspondences (\textit{e.g.}, the class-related attributes), and ii) discards the semantic-unrelated visual tokens of low visual-semantic correspondences (\textit{e.g.}, the meaningless image background) by fusing them into a new token for subsequent learning. Accordingly, ZSLViT progressively learns semantic-related visual representations for desirable visual-semantic interactions in ZSL. As such, ZSLViT achieves significant performances both in seen and unseen classes.
	
	\noindent\textbf{t-SNE Visualization of Visual Features.}
	As shown in Fig. \ref{fig:tsne}, we present the t-SNE visualization of visual features for (a) seen classes and (b) unseen classes on CUB, learned by the CNN backbone  (\textit{e.g.}, ResNet101 \cite{He2016DeepRL}), standard ViT \cite{Touvron2021TrainingDI} and our ZSLViT. Results show the visual features extracted from the CNN Backbone are confused between different classes, while the visual features learned by ViT are high-quality. This intuitively shows ViT is more suitable for ZSL task than CNN backbone. Furthermore, the visual features learned by our ZSLViT are desirable intra-class compactness and inter-class separability, because our ZSLViT discovers the semantic-related visual tokens and discards the semantic-unrelated visual tokens in the whole network according to the visual-semantic correspondences. As such, our ZSLViT significantly improves the ZSL performances over the CNN and ViT backbones.

	\noindent\textbf{Hyper-Parameter Analysis.}
	We perform experiments to investigate the effects of various hyper-parameters on CUB, \textit{i.e.}, embedding coefficient $\gamma$ in Eq. \ref{eq:semantic-embedding}, fusing rate $\kappa$ in ViE, and loss weights $\lambda_{VR}$ and $\lambda_{SR}$ in Eq. \ref{eq:sr-vr}. From the results in Fig. \ref{fig:hyper-para}(a), ZSLViT achieves better performances with a relatively large value of $\gamma$ (\textit{i.e.}, 0.9). The reason is that the semantic information should be progressively embedded into visual features for stable optimization. Fig. \ref{fig:hyper-para}(b) indicates that the performance of ZSLViT drops when the fusing rate $\kappa$ in the visual enhancement module is set to small (\textit{e.g.}, $\kappa<0.6$), because some informative visual tokens are discarded.  The results shown in Fig. \ref{fig:hyper-para}(c) and Fig. \ref{fig:hyper-para}(d) show the better effects of our ZSLViT can be achieved when $\lambda_{VR}$ is larger than $\lambda_{SR}$. Because we mainly aim to enhance semantic information into visual representations for subsequent learning. Overall, ZSLViT is robust to all hyper-parameters and easy to optimize. \vspace{-2mm}
	
	\section{Conclusion}\label{Sec5}
	In this work, we devise a novel zero-shot learning framework, \textit{i.e.}, progressive semantic-guided visual Transformer (ZSLViT), to learn semantic-related visual features for effective visual-semantic interactions. By conducting semantic-embedded token learning and visual enhancement at various stages, our ZSLViT effectively discovers the semantic-related visual features and discards semantic-unrelated visual features according to visual-semantic correspondences. We quantitatively and qualitatively demonstrate that ZSLViT achieves consistent improvements over the current state-of-the-art methods on three ZSL benchmarks, e.g., CUB, SUN, and AWA2.

	\bibliographystyle{ieee_fullname}
	\bibliography{mybibfile}
	
\end{document}